\pdfoutput=1
\documentclass[11pt]{article}

\usepackage{lipsum}  
\usepackage{booktabs}
\usepackage{graphicx}
\usepackage{amsmath}
\usepackage{amssymb}
\usepackage{amsbsy}
\usepackage{xspace}
\usepackage{multirow}
\DeclareMathOperator*{\argmax}{argmax}
\usepackage{url}

\newcommand\BLEU{\textsc{Bleu}\xspace}
\newcommand\TER{\textsc{Ter}\xspace}

\usepackage[]{ACL2023}

\usepackage{times}
\usepackage{latexsym}

\usepackage[T1]{fontenc}
\usepackage[utf8]{inputenc}

\usepackage{microtype}

\usepackage{inconsolata}

\title{On Search Strategies for Document-Level Neural Machine Translation}

\author{
Christian Herold \qquad Hermann Ney \\
Human Language Technology and Pattern Recognition Group \\
Computer Science Department\\
RWTH Aachen University \\
D-52056 Aachen, Germany \\
{\tt \{herold|ney\}@cs.rwth-aachen.de}
}

\begin{document}
\maketitle
\begin{abstract}

Compared to sentence-level systems, document-level neural machine translation (NMT) models produce a more consistent output across a document and are able to better resolve ambiguities within the input.
There are many works on document-level NMT, mostly focusing on modifying the model architecture or training strategy to better accommodate the additional context-input.
On the other hand, in most works, the question on how to perform search with the trained model is scarcely discussed, sometimes not mentioned at all.
In this work, we aim to answer the question how to best utilize a context-aware translation model in decoding.
We start with the most popular document-level NMT approach and compare different decoding schemes, some from the literature and others proposed by us.
In the comparison, we are using both, standard automatic metrics, as well as specific linguistic phenomena on three standard document-level translation benchmarks.
We find that most commonly used decoding strategies perform similar to each other and that higher quality context information has the potential to further improve the translation.
\end{abstract}

\section{Introduction}

Neural machine translation (NMT) \cite{bahdanau2014neural, vaswani2017attention} is widely adopted and produces excellent translations for many domains and language pairs.
However, when these automatic translations are evaluated on the document level, they reveal shortcomings when it comes to consistency in style, entity-translation or correct inference of the gender, among other things \cite{laubli2018has, muller2018large, thai2022exploring}.
Document-level NMT aims to resolve these shortcomings by taking the context of a sentence into account during translation.
There exist many works on the topic of document-level NMT, proposing various changes to the standard transformer \cite{vaswani2017attention} architecture and training criteria to improve context incorporation and consequently translation quality.
However, while the modeling and training aspects are covered in great detail in these works, the exact decoding strategy is often not very clearly described and sometimes not mentioned at all.

In this work, we head out to answer the question, which decoding strategy is most beneficial for document-level NMT systems.
We compare all commonly used strategies, as well as some additional ones, on three standard document-level translation benchmarks.
We find that most of the analyzed decoding strategies perform similar to each other.
Also, higher quality context information can lead to better translations in certain scenarios.

\section{Related Work}

The earliest approaches to document-level NMT simply concatenate consecutive sentences without any further changes to the architecture compared to the sentence-level systems \cite{tiedemann2017neural, agrawal2018contextual}.
Later, some changes were made to the vanilla transformer architecture, like segment embeddings \cite{ma2020simple} or attention masking \cite{zhang-etal-2020-long, petrick2022locality} and a move was made towards translating longer segments \cite{junczys-dowmunt-2019-microsoft, 10.1162/tacl_a_00343, zheng2021towards, bao2021g, sun2022rethinking}.
Other works employ a separate encoder to include the additional context on the source side \cite{jean2017does, bawden2017evaluating, zhang2018improving, voita2018context} or make use of the context in a post-editing fashion \cite{voita2019good, xiong2019modeling}.
Further approaches include the usage of a cache \cite{wang2017exploiting, maruf2018document, tu2018learning} or hierarchical attention networks \cite{miculicich2018document, maruf2019selective, wong2020contextual}.
Recently, several works have concluded that the simple concatenation approach used with the vanilla transformer architecture performs as good - if not better - than more complicated approaches that modify the model structure \cite{sun2022rethinking, majumde2022baseline}.
Since we also observed this in our internal comparisons, we decided to focus on this simple approach for our analysis in this work. 

Several works made the argument that the improvements seen in automatic metric scores for document-level NMT systems are from regularization effects rather than from utilizing the additional context information \cite{kim2019and, li2020does, nguyen2021data}.
In order to better asses the improvements gained by document-level NMT, several targeted test suites have been released \cite{muller2018large, bawden2017evaluating, voita2019good, jwalapuram2019evaluating}.
However, all of these are based on just scoring contrastive examples without actually translating anything.
Recently, \citet{jiang-etal-2022-blonde} and \citet{currey2022mt} have released frameworks that allow to score MT systems on their ability to generate contextually correct translations.\footnote{The framework by \citet{currey2022mt} was not yet made available when we conducted our experiments.}

\section{Search Strategies}
\label{sec:methodology}

Training a document-level NMT system that takes the last $k$ sentences as context is straightforward using the standard concatenation strategy \cite{tiedemann2017neural}.
Given some document level training data $(F_n, E_n), n=1, ..., N$, where $(F_n, E_n)$ denotes the $n$-th source-target sentence pair, during training we optimize the parameters $\Theta$ of the model towards
\begin{equation*}
    \hat{\Theta} = \argmax_{\Theta}\left\{ \sum_n \log p_{\Theta}(E_{n-k}^{n}| F_{n-k}^{n}) \right\}.
\end{equation*}
Here, $E_{n-k}^{n}$ denotes the concatenation of the sentences $E_{n-k}, ..., E_n$.

During search, given a document $F_1^M$, we want to find the best translation $\hat{E}_1^M$ according to the model.
Of course, exact search can not be performed and different works have used different methods to generate a translation:
\begin{description}
    \item[\textit{full segment}]\cite{10.1162/tacl_a_00343, bao2021g, sun2022rethinking}: we split the document into non-overlapping parts $F_1^{k}, F_{k+1}^{2k}, ..., F_{M-k}^{M}$ and translate each part separately using
    \begin{equation}
        \hat{E}_{i-k}^{i} = \argmax_{E_{i-k}^{i}}\left\{p(E_{i-k}^{i}|F_{i-k}^{i}) \right\},
        \label{eq:1}
    \end{equation}
    which is approximated using standard beam search on the token level.
    \item[\textit{last sentence}]\cite{bawden2017evaluating, agrawal2018contextual, zhang-etal-2020-long, petrick2022locality, majumde2022baseline}: we split the document into overlapping parts $..., F_{i-k}^{i}, F_{i-k+1}^{i+1}, ...$ and translate each part separately using Equation \ref{eq:1}. From each translated part we choose only the last sentence to get one translation for every sentence in the document.
    \item[\textit{first sentence}]\cite{zhang-etal-2020-long}: similar to \textit{last sentence}, but from each translated part we choose only the first sentence to get one translation for every sentence in the document.
    \item[\textit{2-pass decoding}]\cite{maruf2018document, maruf2019selective, voita2019good, xiong2019modeling}: we first generate a translation $\tilde{E}_1^M$ of the document using a sentence-level NMT system. Then, the final hypothesis $\hat{E}_i$ for each sentence $F_i$ is created using 
    \begin{equation*}
        \hat{E}_{i} = \argmax_{E_{i}}\left\{p(E_{i}|F_{i-k}^{i}, \tilde{E}_{i-k}^{i-1}) \right\}.
    \end{equation*}
    \item[\textit{doc-trans}]\cite{miculicich2018document, voita2019good, garcia2019context, fernandes2021measuring}: we generate the translation sentence by sentence, meaning
    \begin{equation*}
        \hat{E}_{1} = \argmax_{E_{1}}\left\{p(E_{1}|F_{1}) \right\},
    \end{equation*}
    \begin{equation*}
        \hat{E}_{2} = \argmax_{E_{2}}\left\{p(E_{2}|F_{1}^{2}, \hat{E_1}) \right\},
    \end{equation*}
        \begin{equation*}
        ...
    \end{equation*}
    \item[\textit{doc-trans (beam)}]: similar to \textit{doc-trans}, but instead of keeping just the best context $\hat{E}_{1}^{i-1}$, we keep the top-$h$ candidates and prune them after each step $i$, analogous to beam search on the token level. $h=12$ for all our experiments, the same as our token-level beam-size.
    \item[\textit{cheating}]: this is just used as a tool for analysis. The translation of each sentence $F_i$ is created using the true target reference $\mathring{E}_{1}^{M}$ as context
    \begin{equation*}
        \hat{E}_{i} = \argmax_{E_{i}}\left\{p(E_{i}|F_{i-k}^{i}, \mathring{E}_{i-k}^{i-1}) \right\}.
    \end{equation*}
        \item[\textit{no context}]: this is just used as a tool for analysis. The translation of each sentence $F_i$ is created using no context information at all
    \begin{equation*}
        \hat{E}_{i} = \argmax_{E_{i}}\left\{p(E_{i}|F_{i}) \right\}.
    \end{equation*}
\end{description}

The different search strategies also have a different computational cost associated with them.
The biggest factor regarding the decoding cost is the number of forward passes through the model, specifically the decoder, that we have to do.
We list the computational costs for the different decoding approaches in Table \ref{tab:decoding_cost} under the assumption that the document consists of $N$ sentences with average sentence length $L$ and the model uses $k-1$ sentences as context.
\begin{table}[]
\centering
\begin{tabular}{l|r}
\toprule
 & \multicolumn{1}{c}{Cost} \\ \hline
\textbf{sentence-level} & $\mathcal{O}(NL)$ \\ \hline
\textbf{document-level} &  \\
\textit{full segment} & $\mathcal{O}(NL)$ \\
\textit{last sentence} & $\mathcal{O}(NLk)$ \\
\textit{first sentence} & $\mathcal{O}(NLk)$ \\
\textit{2-pass decoding} & $\mathcal{O}(2NL)$ \\
\textit{doc trans} & $\mathcal{O}(NL)$ \\
\textit{doc trans (beam)} & $\mathcal{O}(NLh)$ \\ \bottomrule
\end{tabular}
\caption{Computational cost of decoding (=number of forward passes through the decoder) for each of the search strategies described above. $h$ denotes the sentence-level beam size.}
\label{tab:decoding_cost}
\end{table}
Please note that the decoding time might follow a different dependence than the cost in the above table, since it heavily depends on the available hardware.
For example, \textit{doc trans} and \textit{doc trans (beam)} might have the same decoding time, if we have enough computational resources available, since the additional computations in \textit{doc trans (beam)} can all be done in parallel.

\section{Experiments}

We perform experiments on three document-level translation benchmarks, called \textbf{NEWS} (En$\to$De), \textbf{TED} (En$\to$It) and \textbf{OS} (En$\to$De).
For the details regarding data conditions and preparation, as well as model training, we refer to Appendix \ref{sec:appendix}.
For the context-aware systems, we concatenate 3 adjacent sentences (i.e. $k=3$) using a special token \texttt{<sep>}.
For the two En$\to$De tasks, we also evaluate the systems on the ContraPro test set \cite{muller2018large}.
Instead of scoring and ranking the contrastive examples in ContraPro, as the authors have originally envisioned, we translate the source side to calculate \BLEU and \TER as well as to score the pronoun translations according to Section \ref{subsec:eval}.
We can not evaluate the \textit{full segment} search strategy on ContraPro, because the sentences are not adjacent since they come from different documents.

\subsection{Evaluating Pronoun Translation}
\label{subsec:eval}

% table 1 moved for formatting reasons
\begin{table}[]
\centering
\begin{tabular}{l|r|r|r}
\toprule
 & \multicolumn{1}{c|}{NEWS} & \multicolumn{1}{c|}{TED} & \multicolumn{1}{c}{OS} \\ \hline
\textbf{sentence-level} &  &  &  \\
ref & 4.61 & 4.15 & 2.97 \\
hyp & 1.63 & 1.56 & 1.48 \\ \hline
\textbf{document-level} &  &  &  \\
ref & 4.46 & 3.96 & 2.70 \\
hyp \textit{no context} & 1.64 & 1.53 & 1.47 \\
hyp \textit{full segment} & 1.62 & 1.50 & 1.41 \\
hyp \textit{last sentence} & 1.61 & 1.48 & 1.42 \\
hyp \textit{first sentence} & 1.63 & 1.52 & 1.48 \\
hyp \textit{2-pass decoding} & 1.68 & 1.49 & 1.48 \\
hyp \textit{doc trans} & 1.67 & 1.48 & 1.42 \\
hyp \textit{doc trans (beam)} & 1.67 & 1.48 & 1.41 \\
hyp \textit{cheating} & 1.69 & 1.53 & 1.56 \\ \bottomrule
\end{tabular}
\caption{Perplexity values on the test set for different search strategies.}
\label{tab:ppls}
\end{table}

As further analysis, we measure how well ambiguous pronouns are handled when translating from English to German.
Regarding gender, the English third-person pronoun \lq{}it\rq{} (and its other forms), can be translated to the German words \lq{}er\rq{}, \lq{}sie\rq{} or \lq{}es\rq{}, depending on which noun it refers to.
On the other hand, ambiguities in the formality come from second-person pronouns.
For example, the English word \lq you\rq can be translated to \lq{}sie\rq{} or \lq{}du\rq{} depending if we are in a formal setting or not.
To report accuracies for pronoun (3 classes: male/female/neuter) and formality (2 classes: formal/informal) translation, we extend the BlonDe metric created by \citet{jiang-etal-2022-blonde}\footnote{Our extension can be found in this fork: \url{https://github.com/christian3141/BlonDe}}.
\begin{table*}[t]
\centering
\begin{tabular}{l|rrrr|rr|rrrr}
\toprule
 & \multicolumn{4}{c|}{NEWS} & \multicolumn{2}{c|}{TED} & \multicolumn{4}{c}{OS} \\ \cline{2-11} 
 & \multicolumn{2}{c|}{test} & \multicolumn{2}{c|}{ConPro} & \multicolumn{2}{c|}{test} & \multicolumn{2}{c|}{test} & \multicolumn{2}{c}{ConPro} \\
 & \multicolumn{1}{c}{\BLEU} & \multicolumn{1}{c|}{\TER} & \multicolumn{1}{c}{\BLEU} & \multicolumn{1}{c|}{\TER} & \multicolumn{1}{c}{\BLEU} & \multicolumn{1}{c|}{\TER} & \multicolumn{1}{c}{\BLEU} & \multicolumn{1}{c|}{\TER} & \multicolumn{1}{c}{\BLEU} & \multicolumn{1}{c}{\TER} \\ \hline
\textbf{sentence-level} &  & \multicolumn{1}{r|}{} &  &  &  &  &  & \multicolumn{1}{r|}{} &  &  \\
\textit{external} & $^{\dagger}$32.3 & \multicolumn{1}{r|}{-} & - & - & $^{\ddagger}$33.4 & - & *37.3 & \multicolumn{1}{r|}{-} & *30.5 & - \\
\textit{ours} & 32.8 & \multicolumn{1}{r|}{49.0} & 18.4 & 65.5 & 34.2 & 46.3 & 37.1 & \multicolumn{1}{r|}{43.8} & 29.7 & 52.8 \\ \hline
\textbf{document-level} &  & \multicolumn{1}{r|}{} &  &  &  &  &  & \multicolumn{1}{r|}{} &  &  \\
\textit{no context} & 33.4 & \multicolumn{1}{r|}{48.5} & 18.6 & 65.6 & 34.0 & 46.7 & 36.9 & \multicolumn{1}{r|}{44.5} & 29.4 & 53.2 \\
\textit{full segment} & 33.4 & \multicolumn{1}{r|}{48.6} & - & - & 34.3 & 46.3 & 38.2 & \multicolumn{1}{r|}{43.9} & - & - \\
\textit{last sentence} & 33.4 & \multicolumn{1}{r|}{48.3} & 19.7 & 63.4 & 34.7 & 45.9 & 37.8 & \multicolumn{1}{r|}{43.9} & 31.4 & 51.5 \\
\textit{first sentence} & 33.4 & \multicolumn{1}{r|}{48.6} & 18.8 & 65.6 & 34.1 & 46.3 & 37.8 & \multicolumn{1}{r|}{44.1} & 29.5 & 53.1 \\
\textit{2-pass decoding} & 32.8 & \multicolumn{1}{r|}{48.6} & 19.5 & 63.9 & 34.5 & 46.2 & 37.4 & \multicolumn{1}{r|}{44.4} & 31.1 & 51.8 \\
\textit{doc trans} & 33.0 & \multicolumn{1}{r|}{48.3} & 19.8 & 63.3 & 34.6 & 46.0 & 37.7 & \multicolumn{1}{r|}{44.0} & 31.4 & 51.3 \\
\textit{doc trans (beam)} & 33.0 & \multicolumn{1}{r|}{48.3} & 19.7 & 63.4 & 34.5 & 46.0 & 38.3 & \multicolumn{1}{r|}{43.8} & 31.3 & 51.5 \\
\textit{cheating} & 32.2 & \multicolumn{1}{r|}{49.4} & 19.3 & 65.1 & 34.1 & 46.6 & 39.6 & \multicolumn{1}{r|}{42.9} & 33.3 & 49.1 \\ \bottomrule
\end{tabular}
\caption{\BLEU and \TER scores (in percent) for the different tasks and decoding strategies. External baselines are from $^{\dagger}$ \citet{kim2019and}, $^{\ddagger}$ \citet{yang2022gtrans} and *\citet{huo2020diving}.}
\label{tab:bleu}
\end{table*}
\begin{table}[t]
\centering
\begin{tabular}{l|r|rr}
\toprule
 & \multicolumn{1}{c|}{NEWS} & \multicolumn{2}{c}{OS} \\ \cline{2-4} 
 & \multicolumn{1}{c|}{ConPro} & \multicolumn{1}{c|}{test} & \multicolumn{1}{c}{ConPro} \\
 & \multicolumn{1}{c|}{gender} & \multicolumn{1}{c|}{style} & \multicolumn{1}{c}{gender} \\ \hline
\textbf{sentence-level} & 45.3 & \multicolumn{1}{r|}{59.4} & 41.4 \\ \hline
\textbf{document-level} &  & \multicolumn{1}{r|}{} &  \\
\textit{no context} & 45.9 & \multicolumn{1}{r|}{59.7} & 42.3 \\
\textit{full segment} & - & \multicolumn{1}{r|}{60.8} & - \\
\textit{last sentence} & 56.1 & \multicolumn{1}{r|}{60.3} & 66.5 \\
\textit{first sentence} & 44.9 & \multicolumn{1}{r|}{59.2} & 43.0 \\
\textit{2-pass decoding} & 55.6 & \multicolumn{1}{r|}{59.9} & 65.1 \\
\textit{doc trans} & 56.1 & \multicolumn{1}{r|}{58.7} & 66.4 \\
\textit{doc trans (beam)} & 56.1 & \multicolumn{1}{r|}{60.6} & 66.3 \\
\textit{cheating} & 63.3 & \multicolumn{1}{r|}{73.2} & 73.7 \\ \bottomrule
\end{tabular}
\caption{F1 scores (in percent) for pronoun translation on different test sets.}
\label{tab:f1}
\end{table}
First, we expand the framework to work for German references, by including German NER and POS taggers\footnote{\url{https://spacy.io/models/de}} as well as including German pronoun mappings.
For the gender category, we mostly follow \citet{jiang-etal-2022-blonde}, but additionally require that a corresponding pronoun must also be present in the source sentence.\footnote{In ContraPro, we find 5011/4085/4817 examples for male/female/neuter respectively. The difference to the 4000/4000/4000 reported by \cite{muller2018large} means that in some cases we count multiple occurrences in a single example.}
For the style category, we take into account examples where a second person pronoun appears in the source sentence, and a corresponding formal or informal pronoun appears in the reference.\footnote{In the OS test set, we count 416 and 605 examples for formal and informal examples respectively.}

\subsection{Perplexities}

% table 2 and 3 moved for formatting reasons

First, we compare the perplexities of the hypotheses from the different search strategies, which are listed in Table \ref{tab:ppls}.
The first thing to note is, that the reference has a much higher perplexity than all hypotheses, which is commonly seen for NMT systems.
All document-level search strategies result in different hypotheses, which however have a similar perplexity score.
Surprisingly, the \textit{cheating} setting generates the worst translation perplexity-wise, even worse than using \textit{no context}.
This might be related to the observation, that the reference has a worse perplexity than any hypothesis, which is rather a modelling error than a search error.

\subsection{Automatic Metrics}

Next, we evaluate the hypotheses based on the common automatic metrics \BLEU and \TER.
The results are shown in Table \ref{tab:bleu}.
The hypotheses created with \textit{no context} seem to have the same quality as the sentence-level baseline.
Surprisingly, the true reference as context does not improve performance on the NEWS and TED test sets.
This indicates that the improvements seen on these test sets for the document-level system might not be related to better context incorporation.
On the contrary, the OS system creates the best hypothesis with the true reference as context.
All the actual decoding strategies give similar performance in terms of \BLEU and \TER with \textit{2-pass decoding} being a little bit behind.
A special case is the \textit{first sentence} strategy, which performs quite well on the standard test sets but poorly on ContraPro.
This is, because ContraPro is designed in a way that the left side context is more important for translation than the right side.

Finally, we analyze the quality of the pronoun translation as discussed in Section \ref{subsec:eval}.
In principle, we could calculate the F1 score for both, gender and formality, on all En$\to$De test sets.
However, we discard the cases where one or more classes have less than 100 examples.
This leaves us with the three test sets depicted in Table \ref{tab:f1}.
As a sanity check, we also report the ContraPro accuracies calculated from scoring the contrastive references as described in \cite{muller2018large}.
They are 48.2/45.8 for sentence-level and 68.2/82.2 for document-level for NEWS/OS respectively.
That means, with just scoring, we overestimate the capabilities of the system, but the trend is still consistent.\footnote{The precision and recall are roughly the same for the F1 scores reported in Table \ref{tab:f1}.}
Using the true reference leads to the best results in all cases.
\textit{no context} and \textit{first sentence} leaves us with sentence-level performance on the gender tasks, while all other decoding strategies perform similarly.
For the formality, none of the methods can significantly outperform the sentence level system, although the \textit{cheating} experiment shows that the system could do better if a better context information is provided.
This might be, because segments of 3 sentences are too short to reliably detect if a setting is formal or informal, without access to the true reference.

\section{Conclusion}

In this work, we analyze decoding strategies for document-level NMT systems.
Using the most popular document-level translation approach, we compare different search strategies found in the literature against methods developed by us.
We find that most of the commonly used decoding strategies result in similar performance, both in terms of common automatic metrics, as well as on specific pronoun evaluation tasks.
Therefore, we conclude that it is important to include the context information during decoding, but the exact way in which to do this is not as important.
Also, we find that the document-level systems could actually profit from higher quality context information, in situations where this context is most relevant for translation.

\section*{Acknowledgements}
This work was partially supported by the project HYKIST funded by the German Federal Ministry of Health on the basis of a decision of the German Federal Parliament (Bundestag) under funding ID ZMVI1-2520DAT04A, and by NeuroSys which, as part of the initiative “Clusters4Future”, is funded by the Federal Ministry of Education and Research BMBF (03ZU1106DA).

\section*{Limitations}
In this work, we limit our experiments to the most commonly used document-level system architecture and training criterion.
Other approaches exist, which might exhibit a different behavior in decoding.
Two out of the three document-level translation tasks we use in this work are low resource with less than 500k sentence-pairs as training data.
We chose these tasks due to computational limitations and to be better comparable to other works, but higher resource scenarios are more realistic for actual applications.
We limit the analysis of pronoun translation to the English-German language pair.
Also, there are other aspects of document-level NMT, like consistent translation of entities, which we did not consider in our analysis.

% \section*{Acknowledgements}

\bibliography{anthology,custom}
\bibliographystyle{acl_natbib}
% Limitations
% \appendix

% \section{Example Appendix}
% \label{sec:appendix}

% This is a section in the appendix.

\clearpage

\appendix

\section{Appendix}
\label{sec:appendix}

For the \textbf{NEWS En$\to$De} task, the parallel training data (around 300k sentence pairs, news-domain) comes from the \texttt{NewsCommentaryV14} corpus\footnote{\url{https://data.statmt.org/news-commentary/v14/}}.
As validation/test set we use the WMT \texttt{newstest2015}/\texttt{newstest2018} test sets from the WMT news translation tasks \cite{farhad2021findings}.
For the \textbf{TED En$\to$It} task, the parallel training data (around 200k sentence pairs, scientific-talks-domain) comes from the IWSLT17 Multilingual Task \cite{cettolo2017overview}.
As validation set we use the concatenation of \texttt{IWSLT17.TED.dev2010} and \texttt{IWSLT17.TED.tst2010} and as test set we use \texttt{IWSLT17.TED.tst2017.mltlng}.
For the \textbf{OS En$\to$De} task, the parallel training data (around 22.5M sentence pairs, subtitle-domain) comes from the \texttt{OpenSubtitlesV2018} corpus \cite{lison2018opensubtitles2018}.
We use the same train/validation/test splits as \citet{huo2020diving} and additionally remove all segments that are used in the ContraPro test suite \cite{muller2018large} from the training data.
The data statistics for all tasks can be found in Table \ref{tab:data}.

\begin{table}[h!]
\centering
\begin{tabular}{l|l|r|r}
\toprule
\multicolumn{1}{c|}{task} & \multicolumn{1}{c|}{dataset} & \multicolumn{1}{c|}{\# sent.} & \multicolumn{1}{c}{\# doc.} \\ \hline
NEWS & train & 330k & 8.5k \\
 & valid & 2.2k & 81 \\
 & test & 3k & 122 \\
 & ContraPro & 12k & 12k \\ \hline
TED & train & 232k & 1.9k \\
 & valid & 2.5k & 19 \\
 & test & 1.1k & 10 \\ \hline
OS & train & 22.5M & 29.9k \\
 & valid & 3.5k & 5 \\
 & test & 3.8k & 5 \\
 & ContraPro & 12k & 12k \\ \bottomrule
\end{tabular}
\caption{Data statistics for the different document-level translation tasks.}
\label{tab:data}
\end{table}

Since in the original release of ContraPro only left side context is provided, we extract the right side context ourselves from \texttt{OpenSubtitlesV2018} based on the meta-information of the segments.

We tokenize the data using byte-pair-encoding \cite{sennrich2016neural, DBLP:conf/acl/Kudo18} with 15k joint merge operations (32k for OS En$\to$De).
The models are implemented using the fairseq toolkit \cite{ott2019fairseq} following the transformer base architecture \cite{vaswani2017attention} with dropout 0.3 and label-smoothing 0.2 for \textbf{NEWS En$\to$De} and \textbf{TED En$\to$It} and dropout 0.1 and label-smoothing 0.1 for \textbf{OS En$\to$De}.
This resulted in models with ca. 51M parameters for NEWS and TED and ca. 60M parameters for OS for both the sentence-level and the document-level systems.
All systems are trained until the validation perplexity does no longer improve and the best checkpoint is selected using validation perplexity as well.
Training took around 24h for NEWS and TED and around 96h for OS on a single NVIDIA GeForce RTX 2080 Ti graphics card.
Due to computational limitations, we report results only for a single run.
For the generation of segments (see Section \ref{sec:methodology}), we use beam-search on the token level with beam-size 12 and length normalization.
To calculate \BLEU \cite{papineni2002bleu} and \TER \cite{snover2006study} we use SacreBLEU \cite{post2018call}.

\end{document}